\documentclass[journal]{IEEEtran}
\def\etal{\textit{et al}}
\usepackage{color}

\usepackage{cite}

\ifCLASSINFOpdf
  \usepackage[pdftex]{graphicx}
  \graphicspath{ {images/} }
\else
\fi
\usepackage{amsmath, amssymb}
\usepackage{url}

\begin{document}
%
\title{Skepxels: Spatio-temporal Image Representation of Human Skeleton Joints for Action Recognition}
%
%
%

\author{Jian~Liu,
        Naveed~Akhtar,
        and~Ajmal~Mian 
\thanks{J. Liu, N. Akhtar and A. Mian are with the School of Computer Science and Software Engineering, The University of Western Australia, Perth,
WA, 6009 Australia e-mail: (jian.liu@research.uwa.edu.au, naveed.akhtar@uwa.edu.au and ajmal.mian@uwa.edu.au).}
}

%
%

\markboth{Journal of \LaTeX\ Class Files,~Vol.~14, No.~8, August~2015}%
{Shell \MakeLowercase{\textit{et al.}}: Bare Demo of IEEEtran.cls for IEEE Journals}
%



\maketitle

\begin{abstract}
Human skeleton joints are popular for action analysis since they can be easily extracted from videos to discard background noises. However, current skeleton representations do not fully benefit from machine learning with Convolutional Neural Networks (CNNs). We propose ``Skepxels'' a spatio-temporal representation for skeleton sequences to fully exploit the correlations between joints using the kernels of CNNs. We transform skeleton videos into images of flexible dimensions using Skepxels and develop a CNN-based framework for effective human action recognition using the resulting images. Skepxels encode rich spatio-temporal information about the skeleton joints in the frames by maximizing a unique distance metric, defined collaboratively over the distinct joint arrangements used in the skeletal images. Moreover, they are flexible in encoding compound semantic notions such as location and speed of the joints. The proposed action recognition exploits the representation in a hierarchical manner by first capturing the micro-temporal relations between the skeleton joints with  the Skepxels and then exploiting their macro-temporal relations by computing the Fourier Temporal Pyramids over the CNN features of the skeletal images. We extend the Inception-ResNet CNN architecture with the proposed method and improve the state-of-the-art accuracy by $4.4\%$ on the large scale NTU human activity dataset. On the medium-sized NUCLA and UTD-MHAD datasets, our method outperforms the existing results by $5.7\%$ and $9.3\%$ respectively.
\end{abstract}


%
\IEEEpeerreviewmaketitle

\section{Introduction}
\label{sec:Intro}

\IEEEPARstart{H}{uman} action recognition~\cite{6409441, 7042326, 6894548} is a long standing problem in Computer Vision, with applications to smart surveillance, video retrieval, and human-computer interaction etc.
Within the domain of video-based action recognition, extracting human skeleton joints from videos to perform this task is a popular technique as it is robust to the variations in clothing, illumination condition and background~\cite{devanne20153, yang2017latent,poppe2010survey,herath2017going,xia2012view,vemulapalli2014human,du2015hierarchical,vemulapalli2016rolling,wang2017modeling}. Moreover, recent methods can extract skeleton data in real-time from single view RGB videos~\cite{mehta2017vnect}. Convolutional Neural Networks (CNNs)~\cite{AlexNet,GoogLeNet,ResNet} are popular for processing raw images~\cite{wang2015action,HPM+TM,karpathy2014large,tran2015learning} because they effectively exploit the correlation between the local pixels in the images, which is the key to accurate image classification. We envisage that higher human action recognition accuracy can be achieved analogously by capitalizing on the correlations between the skeleton joints. This is possible by arranging the skeleton joints in images and allowing  CNNs to be directly trained on such images. However, the low number of joints and the inherent dissimilarity between the skeletons and images restrict the utility of CNNs for processing the skeleton data.  A major motivation behind this work is to fully exploit the perpetual advances in CNNs for skeletal action data. We demonstrate that CNNs can lead to state-of-the-art action recognition performance using skeletal time-series data alone under the appropriate representation proposed in this work.

Previous attempts~\cite{ke2017new, du2015skeleton} of using CNNs for human skeleton data generally use the skeleton joints of a video frame to form an image column. This severely limits the number of joints per frame in the receptive field of the 2D kernels of the CNNs, restricting the kernel's capacity to exploit the correlations between multiple skeleton joints. These methods also find it inevitable to up-sample the skeleton data to construct appropriate size images for using 
pre-trained networks. On one hand, up-sampled images suffer from ill-defined semantics; on the other, the image generation process adds noise to the data. Finally, existing methods do not take into account the different ways in which skeleton joints can be arranged to form an image.

We propose an atomic visual unit \emph{Skepxel} - skeleton picture element or skeleton pixel, to construct skeletal images of flexible dimensions that can be directly processed by modern CNN architectures without
any re-sampling. Skepxels are constructed by organizing a set of distinct skeleton joint arrangements from multiple frames into a single tensor. The set is chosen under a unique distance metric that is collectively defined over the  joint arrangements for each frame. Unlike previous works where skeleton joints of a frame were arranged in a column, we arrange them in a 2D grid to take full advantage of the 2D kernels in CNNs. 
The temporal evolution of the joints is captured by employing Skepxels from multiple frames into one image.
Thus, the resulting image is a compact representation of rich spatio-temporal information about the action. 
Owing to the systematic construction of the skeletal images, it is also possible to encode multiple semantic notions about the joints in a single image - shown by encoding ``location'' and ``velocity'' of the joints.

We also contribute a framework that uses the proposed Skepxels representation for human action recognition.
To that end, we hierarchically capture the micro-temporal relations between the joints in the frames using Skepxels and exploit the macro-temporal relations between the frames by computing the Fourier Temporal Pyramids~\cite{wang2012mining} of the CNN features of the skeletal images.
We demonstrate the use of skeletal images of different sizes with the Inception-ResNet~\cite{szegedy2017inception}.
Moreover, we also enhance the network architecture for the proposed framework.
The proposed technique is thoroughly evaluated using the NTU Human Activity Dataset~\cite{shahroudy2016ntu}, Northwestern-UCLA Multiview Dataset~\cite{AOG} and UTD Multimodal Human Action Dataset~\cite{chen2015utd}.
Our approach improves the state-of-the-art performance on the large scale dataset~\cite{shahroudy2016ntu} by $4.4\%$, whereas the accuracy gain on the remaining two datasets is $5.7\%$ and $9.3\%$.

This article is organized as follows. In Section~\ref{sec:RW}, we review the related literature. The proposed approach is discussed in Section~\ref{sec:PA}. We introduce the datasets used in our evaluation in Section~\ref{sec:DS}, and discuss the experiments in Section~\ref{sec:Exp}. For a thorough analysis, we provide further ablation studies in Section~\ref{sec:AbS}. The paper concludes in Section~\ref{sec:Conc}.
 

\section{Related Work}
\label{sec:RW}
With the easy availability of reliable human skeleton data from RGB-D sensors, the use of skeleton information in human action recognition  is becoming very popular. Skeleton based action analysis is becoming even more promising because of the possibility of extracting skeleton data in real time using a single RGB camera~\cite{mehta2017vnect}. 
Skeleton data can be directly used to recognize human actions. For instance, Devanne~\etal~\cite{devanne20153} represented the 3-D coordinates of skeleton joints and their change over time as trajectories, and formulated the action recognition problem as computing the similarity between the shape of trajectories in a Riemannian manifold. The joint trajectories model the temporal dynamics of actions, and remain invariant to geometric transformation. Yang~\etal~\cite{yang2017latent} proposed a mid-level granularity of joints called skelets, which can be used to describe the intrinsic interdependencies between skeleton joints and action classes. The authors  integrated multi-task learning and max-margin learning to capture the correlation between skelets and action classes. To balance the skelet-wise relevance and action-wise relevance, a joint structured sparsity inducing regularization is also integrated into their framework. 

Skeleton information is also commonly used in guiding the action representation in other image and video modalities. Cao~\etal~\cite{cao2018body} used extracted body joints to guide the selection of convolutional layer activations of RGB input action videos. They pooled the activations of 3-D convolutional feature maps according to the position of body joints, and thus created discriminative spatio-temporal video descriptors for action recognition. To facilitate end-to-end training, they proposed a two-stream framework with bilinear pooling, with one stream extracting visual features and the other locating key-points of the features maps.

Zanfir \etal~\cite{zanfir2013moving} proposed a moving pose descriptor which considers both pose information and the differential quantities of the skeleton joints for human action recognition.  Their approach is non-parametric and therefore can be used with small amount of training data or even with one-shot training.
Du \etal \cite{du2015skeleton} transformed the skeleton sequences into images by concatenating the joint coordinates as vectors and arranged these vectors in a chronological order as columns of an image. The generated images are resized and passed through a series of adaptive filter banks. Although effective, their approach is based on global spatial and temporal information and fails to exploit the local correlation of joints in skeletons. In contrast, our proposed approach models the global and local temporal variations simultaneously.

Veeriah \etal \cite{veeriah2015differential} proposed to use a differential Recurrent Neural Network (dRNN) to learn the salient spatio-temporal structure in a skeleton action. They used the notion of ``Derivative of States'' to quantify the information gain caused by the salient motions between the successive frames, which guides the dRNN to gate the information that is memorized through time. Their method relies on concatenating 5 types of hand-crafted skeleton features to train the proposed network.
Similarly, Du \etal \cite{du2015hierarchical} applied a hierarchical RNN to model skeleton actions. They divided the human skeleton into five parts according to human physical structure. Each part is fed into a bi-directional RNN and the outputs are hierarchically fused for the higher layers.  One potential limitation of this approach is that the definition of body part is dataset-specific, which causes extra preprocessing when applied to different action datasets.

Shahroudy \etal \cite{shahroudy2016multimodal} also used the division of body parts and proposed a multimodal-multipart learning method to represent the dynamics and appearance of body. They selected the discriminative body parts by integrating a part selection process into their learning framework. In addition to the skeleton based features, they also used hand-crafted features for depth modality, such as LOP (local occupancy patterns) and local HON4D (histogram of oriented 4D normals) around each body joint.
Vemulapalli and Chellappa \cite{vemulapalli2016rolling} represented skeletons using the relative 3D rotations between various body parts. They applied concept of rolling maps to model skeletons as points in the Lie group, and then modeled human actions as curves in the Lie group. By combing the logarithm map with the rolling maps, they managed to unwrap the action curves and performed classification in the non-Euclidean space. 

Based on the intuition that the traditional Lie group features may be too shallow to learn a robust recognition algorithm for skeleton data, Huang \etal \cite{huang2016deep} incorporated the Lie group structure into deep learning, to transform the high-dimensional Lie group trajectory into temporally aligned Lie group features for skeleton-based action recognition. Their learning structure (LieNet) generalizes the traditional neural network model to non-Euclidean Lie groups.  One issue  with LieNet is that it is mainly designed to learn spatial features of skeleton data, and does not take full advantage of the rich temporal cues of human actions. To leverage both spatial and temporal information in skeleton sequences, Kerola \etal \cite{kerola2017cross} used a novel graph representation to model skeletons and keypoints as a temporal sequence of graphs, and  applied the spectral graph wavelet transform to create the action descriptors. By carefully selecting interest points when building the graphs, their approach achieves view-invariance and is able to capture human-object interaction.

Ke \etal \cite{ke2017new} transformed a skeleton sequence into three clips of gray-scale images. Each clip consists of four images, which encode the spatial relationship between the joints by inserting reference joints into the arranged joint chains. They employed the pre-trained VGG19 model to extract image features and applied the temporal mean pooling to represent an action. 
A Multi-Task Learning was proposed for classification.
Wang and Wang~\cite{wang2017modeling} proposed a two-stream RNN architecture to simultaneously exploit the spatial relationship of joints and temporal dynamics of the skeleton sequences. In the spatial RNN stream, they used a chain-like sequence and a traversal sequence to model the spatial dependency, which restricts modeling all possibilities of the joint movements.

Kim and Reiter \cite{kim2017interpretable} proposed a Res-TCN architecture to learn spatial-temporal representation for skeleton actions. They constructed per-frame inputs to the Res-TCN by flatting 3D coordinates of the joints and concatenating values for all the joints in a skeleton. Their method improves interpretability for skeleton action data, however, it does not effectively leverage the rich spatio-temporal relationships between different body joints.
To better represent the structure of skeleton data, Liu \etal \cite{liu2016spatio} proposed a tree traversal algorithm to take the adjacency graph of the body joints into  account. They processed the joints in top-down and bottom-up directions to keep the contextual information from both the descendants and the ancestors of the joints. Although this traversal algorithm discovers spatial dependency patterns, it has the limitation that the dependency of joints from different tree branches can not be easily modeled.

\section{Proposed Approach}
\label{sec:PA}

Restricted by the small number of joints in a human skeleton, existing approaches for converting the skeleton data into images generally result in smaller size images than what is required for the mainstream CNN architectures e.g.~VGG~\cite{simonyan2014very}, Inception~\cite{szegedy2015going}, ResNet~\cite{he2016deep}. Consequently, the images are up-sampled to fit the desired network architectures \cite{du2015skeleton,ke2017new} which  imports unnecessary noise in the data. 
This also compromises the effectiveness of the network kernels that are unable to operate on physically meaningful discrete joints. 
One potential solution is to design new CNN architectures that are better suited to  the smaller images. 
However, small input image size restricts the receptive fields of the convolution  kernels as well as the network depth. 
As a result, the network may not be able to appropriately model the skeleton data. 

In this paper, we address this problem by mapping the skeleton data from a fixed length sequence to an image with the help of a basic building block (similar to pixel). The resulting image is rich in both spatial and temporal information of the skeleton sequences, and can be constructed to match arbitrary input dimensions of the existing network architectures. The approach is explained below.


\subsection{Skeleton Picture Elements (Skepxels)}
\begin{figure}[t]
\centering
\includegraphics[width=0.45\textwidth]{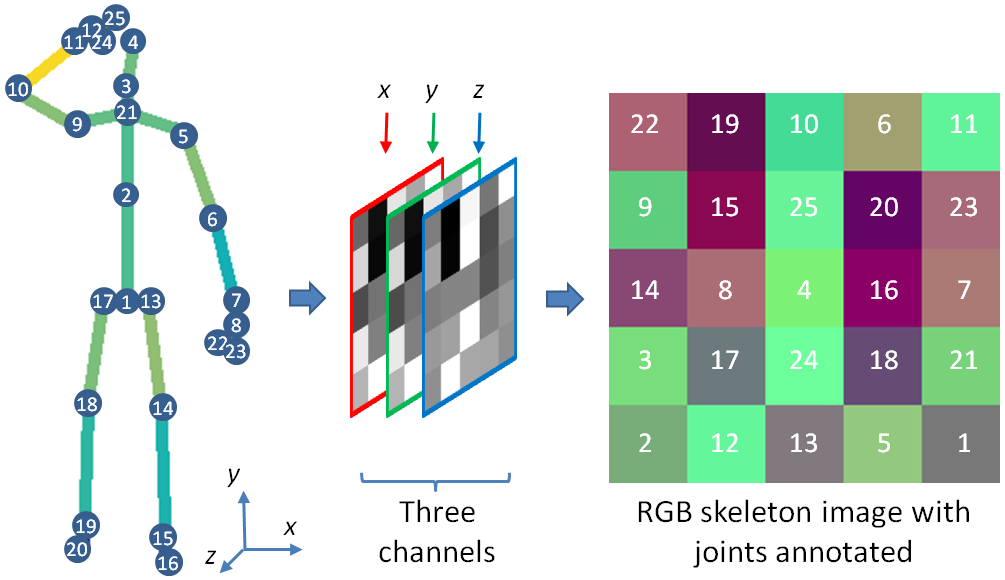}
\caption{Illustration of a \emph{Skepxel} rendered as an RGB image patch. The numbers on skeleton and color image share the joint description. e.g.\ 3-neck, 18-right knee, 21-spine, etc.}
\label{fig:basic_element}
\end{figure}

We propose to map a skeleton sequence to an image ${\bf I} \in \mathbb R^{H\times W \times 3}$ with the help of \emph{Skepxels}.
A Skepxel is a tensor $\boldsymbol\psi \in \mathbb R^{h \times w \times 3}$ obtained by arranging the indices of  the skeleton joints in a 2D-grid and  encoding their coordinate values along the third dimension. 
We treat the skeleton in a video as a set $\mathcal S \subseteq \mathbb R^3$ such that its $j^{\text{th}}$ element, i.e.~${\bf s}_j \in \mathbb R^3$ represents the Cartesian coordinates of the $j^{\text{th}}$ skeleton joint.
Thus, the cardinality of $\mathcal S$, i.e.~$|\mathcal S| \in \mathbb R$ denotes the total number of joints in the skeleton.
For $\boldsymbol\psi$, it entails $h \times w = |\mathcal S|$.
This formulation allows us to represent a Skepxel as a three-channel image patch, as illustrated in Fig.~\ref{fig:basic_element}.
We eventually construct the image ${\bf I}$ by concatenating multiple Skepxels for a  skeleton sequence.  

Owing to the square shaped kernels of CNN architectures, the skeletal information in images is likely to be processed more effectively for square/near square shaped building blocks of the images.
Therefore, our representation  constrains the height and the width of the Skepxels to be as similar as possible.

\subsection{Compact spatial coding with Skepxels} \label{spatial_extension}
A Skepxel constructed for a given skeleton frame encodes the spatial locations of the skeleton joints.
Considering the convolution operations involved in  CNN learning, it is apparent that different arrangements of the joints in a Skepxel can  result in a different behavior of the models. 
This is fortuitous, as we can encode more information in the image $\bf I$ for the CNNs by constructing it with multiple Skepxels that employ different joint arrangements. However, the image must use only a few (but highly relevant) Skepxels for keeping the representation of  the skeleton sequence compact.

Let $\mathcal{A} \subseteq \mathbb R^{h \times w}$ be a set of 2D-arrays, with its $i^{\text{th}}$ element ${\bf A}_i \in \mathbb R^{h \times w}$ representing the $i^{\text{th}}$ possible arrangement of the skeleton joints for a Skepxel.
The cardinality of this set can be given as $|\mathcal{A} | = (h \times w)!$.
Even for a video containing only a $25$-joint skeleton, the total number of possible arrangements of the joints for a Skepxel is $\sim 1.55\times10^{25}$.
Assume that we wish to use only $m$ Skepxels in  ${\bf I}$ for the sake of  compactness, we must then select the joint arrangements for those Skepxels from a possible $^{|\mathcal A|}C_{m}$ combinations, which becomes a prohibitively large number for the practical cases (e.g. $^{(4\times 4)!}C_{16} > 10^{199}$). Therefore, a principled approach is required to choose the suitable arrangements of the joints to form the desired Skepxels.

\begin{figure}[t]
\centering
\includegraphics[width=0.4\textwidth]{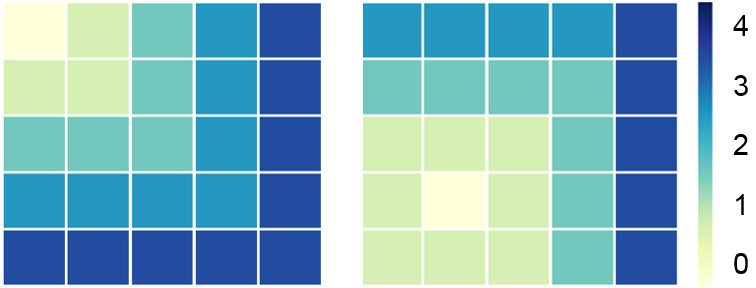}
\caption{Illustration of the employed definition of the radial distance on $5 \times 5$ grids. If the joint $\alpha_i$ is located at [1,1] position in ${\bf A}_j^m$, the left $5 \times 5$ grid is used. For the joint location [4,2],  the right grid is used. There are 25 such grids in total to measure the distance of skeleton joints among $m$ arrangements.}
\label{fig:distance_metrics}
\end{figure}


To select the $m$ arrangements for the same number of Skepxels, we define a metric $\Delta (\mathcal A^m)\! \rightarrow \!\gamma$ over an arbitrary subset $\mathcal A^m$ of $\mathcal A$, where $|\mathcal A^m|\!=\! m$, such that
\begin{align}
\Delta (\mathcal A^m) = \sum\limits_{j = 1}^{|\mathcal A^m|} \sum\limits_{i=1}^{|\mathcal S|}\delta(\alpha_i, {\bf A}_j^m).
\label{eq:metric}
\end{align}
In Eq.~(\ref{eq:metric}), ${\bf A}_j^m$ denotes the $j^{\text{th}}$ element of $\mathcal A^m$ and $\alpha_i$ is  the $i^{\text{th}}$ element of the set $\{1,2,...,|\mathcal S|\}$.
The function $\delta(.,.)$ computes the cumulative radial distance between the location of the joint $\alpha_i$ in ${\bf A}_j^m$ and its locations in the remaining elements of $\mathcal A^m$.
 Let $(x,y)$ denote the indices of $\alpha_i$ in ${\bf A}^m_j$, and $(x_q,y_q)$ denote its indices in any other ${\bf A}_q^m \in \mathcal A^m$, then $\delta(\alpha_i, {\bf A}_j^m) = \sum_{q\neq j, q = 1}^{|\mathcal A^m|-1} \text{max}(abs(x-x_q), abs(y-y_q))$, where $abs(.)$ computes the absolute value.
As per the definition of  $\Delta(.)$, $\gamma$ is a distance metric defined over a set of $m$ possible arrangements of the skeleton joints such that a higher value of $\gamma$ implies a better scattering of the joints in the considered $m$ arrangements.
The  notion of the radial distance used in Eq.~(\ref{eq:metric}) is illustrated in Fig.~\ref{fig:distance_metrics}.
Noticing the image patterns in the figure, we can see the relevance of this metric  for the CNNs that employ square shaped kernels, as compared to the other metrics, e.g.~Manhattan distance.


Due to better scattering,  the skeleton joint arrangements with the larger $\gamma$ values are generally preferred by the CNN architectures to achieve higher accuracy. 
Moreover, different sets of arrangements with similar $\gamma$ values were found to  achieve similar accuracies.
Interestingly, this implies  that for the CNNs the relative positions of the joints in the Skepxels become more important as compared to their absolute positions.
This observation preempts us to construct Skepxels with the skeleton joint arrangements based on the semantics of the joints. On the other hand, selection of the best set of arrangements from the  $^{|\mathcal A|}C_{m}$ possibilities is an NP-hard problem for all practical cases.

\begin{figure}[t]
\centering
\includegraphics[width=0.4\textwidth]{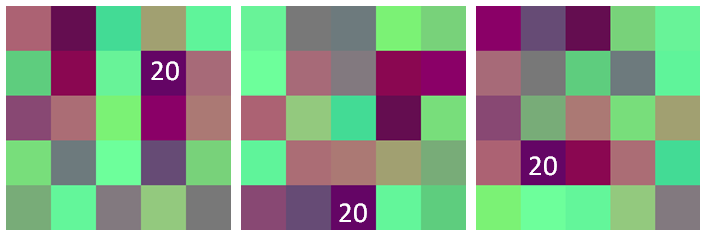}
\caption{A group of \emph{Skepxels} generated for a single skeleton frame. The same color corresponds to the same joint. Only joint number 20 is marked for better visibility.}
\label{fig:different_arrangments}
\end{figure}

We devise a pragmatic strategy to find a suitable set of the skeleton joint arrangements for the desired $m$ Skepxels. 
That is, we empirically choose a threshold $\gamma_t$ for the Skepxels and generate $m$ matrices in $\mathbb R^{h \times w}$ such that the coefficients of the matrices are sampled uniformly at random in the range $[1, h\times w]$, without replacement.
We consider these matrices as the elements of $\mathcal A^m$ if their $\gamma$ value is larger than $\gamma_t$. We use the resulting $\mathcal A^m$ to construct the $m$ Skepxels.
The Skepxels thus created encode a largely varied skeleton joint arrangements in a compact manner.  Fig.~\ref{fig:different_arrangments}, illustrates three Skepxels created by the proposed scheme for a single skeleton frame containing 25 joints. The Skepxels are shown as RGB image patches. 
In our approach, we let $m = H/h$ and construct a tensor $\boldsymbol\Psi \in \mathbb R^{H\times w \times 3}$ by the row-concatenation of the Skepxels $\boldsymbol\psi_{i \in \{1,2,...,m\}}$.
The constructed tensor $\boldsymbol\Psi$ is rich in the spatial information of the joints in a single frame of the video.

\begin{figure}[t]
\includegraphics[width=0.4\textwidth]{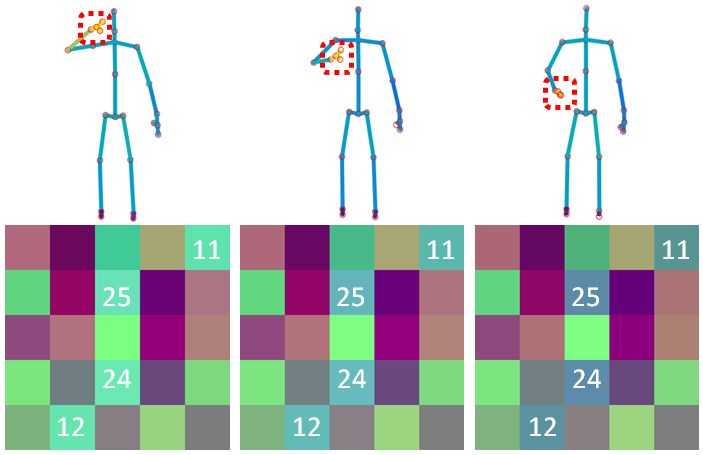}
\centering
\caption{Illustration of  \emph{Skepxels} for a sequence of frames. The same location corresponds to the same joint. Only three Skepxels for three frames are shown.}
\label{fig:different_frames}
\end{figure}

\subsection{Compact temporal coding with Skepxels} \label{temporal_extension}
\begin{figure}[t]
\centering
\includegraphics[width=0.45\textwidth]{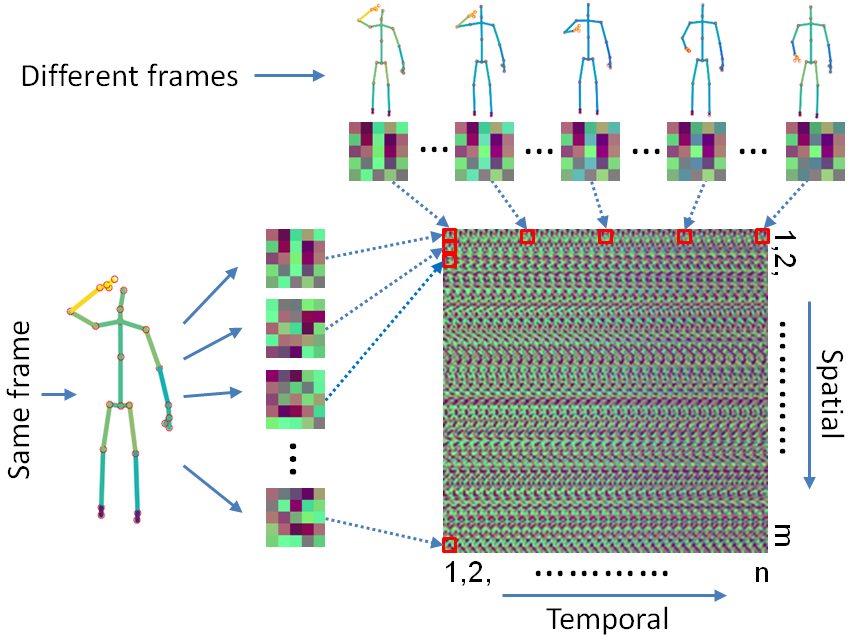}
\caption{The final image is compactly constructed with the \emph{Skepxels} along spatial and temporal dimensions.}
\label{fig:image_generation}
\end{figure}

To account for the temporal dimension in a sequence of the skeleton frames, we compute  the tensor $\boldsymbol\Psi_i$ for the $i^{\text{th}}$  frame in the $n$-frame sequence and  concatenate those tensors in a column-wise manner to construct the desired image ${\bf I}$. 
As illustrated in Fig.~\ref{fig:different_frames}, for a sequence of frames the appearance of a Skepxel  changes specifically at the locations of the active  joints for the action - indicating effective encoding of the action dynamics by Skepxels.
The concatenation of $\boldsymbol\Psi_{i \in \{1,2,...,n\}}$ ensures that the dynamics are recorded in  ${\bf I}$ under $m$ suitable Skepxels, making the representation spatially and temporally rich.   
The formation of the final image by concatenating $\boldsymbol\Psi_i, \forall i$  is illustrated in Fig.~\ref{fig:image_generation}. 

Different action videos may contain various number of skeletal frames. 
For the videos that comprise the skeleton sequences with more than $n$ frames, we create multiple images from the same video and label them according to the  action label.
For the videos with fewer than $n$ frames, we found that the simple strategy of interpolating between the frames works well to construct the image of the desired size.
Note that, the images resulting from the proposed method capture the temporal dynamics in the  raw skeleton data. 
By fixing the length of the temporal window to $n$, the images are able to encode the micro-temporal movements that are expected to model the fine motion patterns contributing  to the classification of the entire action video. In Section~\ref{sec:TEC}, we also discuss the exploitation of the macro-temporal relationships with the proposed representation.

\subsection{Modeling joint speed with Skepxels}
\label{sec:Speed}
The modular approach to construct  images with  the Skepxels not only allows us to easily match the input dimensions of an existing CNN architecture, it also provides the flexibility to encode a notion that is semantically different than the ``locations'' of the skeleton joints.
We exploit this fact to extend our representation to the skeleton joint ``speeds'' in the frame sequences.
To that end, we construct the Skepxels similar to the procedure described above, however instead of using the Cartesian coordinate values for the joints we use the differences of these values for the same joints in the consecutive frames.
A Skepxel thus created encodes the speeds of the joint movements, where  the time unit is  governed by the video frame-rate.  We refer to the final tensors constructed with the joint coordinates as the \emph{location} images, and the tensors constructed using the joint speeds as the \emph{velocity} images.

\begin{figure}[t] 
\centering
\includegraphics[width=0.4\textwidth]{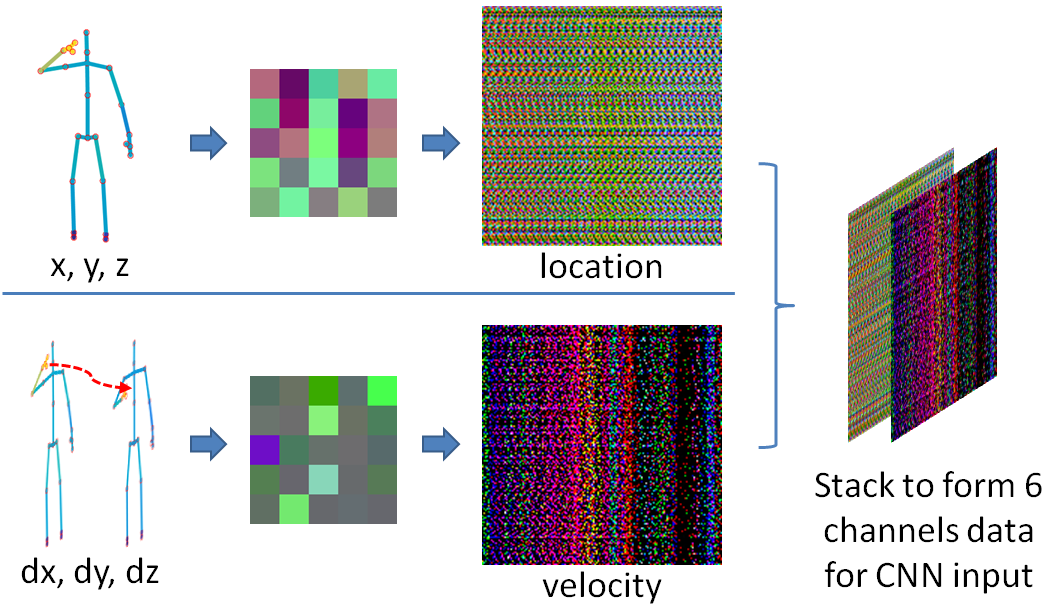}
\caption{Joint differences between the consecutive skeleton frames are calculated to construct the \emph{velocity} images, which are appended to  the \emph{location} images.}
\label{fig:target_images}
 \end{figure}

For many actions, the speed variations among different skeleton joints is an important cue for distinguishing between them (e.g.~{\fontfamily{qcr}\selectfont walking} and {\fontfamily{qcr}\selectfont  running}), and it is almost always supplementary to the information encoded in the absolute locations of the joints.
Therefore, in our representation, we augment the final image by appending the three speed channels $dx, dy, dz$ to the  three location channels $x, y, z$. This augmentation is illustrated in Fig.~\ref{fig:target_images}.  
We note that unless allowed by the  CNN architecture under consideration, the augmentation with the speed channels is not mandatory in our representation.
Nevertheless, it is  desirable for better action recognition accuracy, which will become evident from  our experiments in Section~\ref{sec:Exp}.

\subsection{Normalization and data augmentation}
\label{sec:NAD}
Before converting the skeleton data into images using the proposed method, we perform  the normalization of the raw skeleton data. 
To do so, we anchor the {\fontfamily{qcr}\selectfont hip} joint in a skeleton to the origin of the used Cartesian coordinates, and align the virtual vector between the  {\fontfamily{qcr}\selectfont left-shoulder} and the {\fontfamily{qcr}\selectfont right-shoulder} of the skeleton to the x-axis of the coordinate system. 
This normalization strategy also results in mitigating the translation and viewpoint variation effects in the skeleton data by filtering out the motion-irrelevant noises.  
A further normalization is performed over the channels of the resulting images to restrict the values of the pixels in the range $[0, 255]$. Both types of normalizations are carried out on the training as well as the testing data.

In order to augment the data, we make use of the additive Gaussian noise. We draw samples from the zero Mean  Gaussian distribution with 0.02 Standard Deviation and add those samples to  the skeleton joints in the frame sequences to double the training data size.
This augmentation strategy is based on the observation that slight variations in the joint locations/speeds generally do not vary the skeletal information significantly enough to change the label of the associated action.  
For our experiments, doubling the training data size  already resulted in a significant performance gain over the existing approaches. Therefore, no further data augmentation was deemed necessary for the experiments.

\begin{figure*}[t]
\centering
\includegraphics[width=0.75\textwidth]{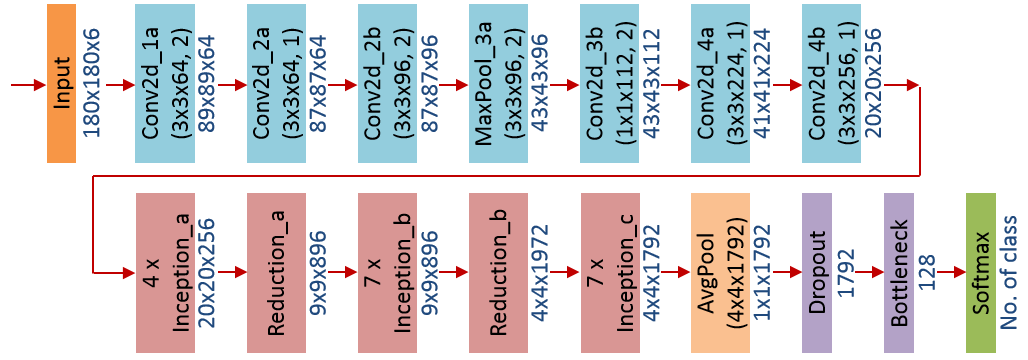}
\caption{Modified architecture of the Inception-ResNet \cite{szegedy2017inception}: The ``STEM'' part of the network is extended to fit the augmented 6-channel input images. The input and output sizes are described as $rows \times columns \times channels$. The kernel is specified as $rows \times columns \times filters, stride$.}
\label{fig:extended_inception_resnet}
\end{figure*}

\subsection{Processing skeletal images with CNNs} \label{model_training}
Due to its flexibility, the proposed mapping of the skeletal information  to the  image-like tensors allows us to exploit a wide variety of existing (and potentially future) CNN architectures to effectively process the information in the skeleton frame sequences. 
To demonstrate this, we employ the  Inception-ResNet \cite{szegedy2017inception} as the test bed for our representation.
This recent CNN architecture has been successful in the general image classification task \cite{deng2009imagenet}, as well as the specific tasks such as face recognition~\cite{schroff2015facenet}.
More importantly, the architecture allows for a variable input image size both in terms of the spatial dimensions and the number of color channels of the image.

First, we trained the Inception-ResNet from scratch by constructing the skeletal images of different dimensions (without the speed channel augmentation).
This training resulted in a competitive performance of the network for a variety of image sizes - details provided in Section~\ref{sec:Exp}.
We strictly followed the original work~\cite{szegedy2017inception} for the training methodology, which demonstrates the compatibility of the proposed representation with the existing frameworks.
In our experiments, training the network from scratch was consistently  found to be more effective than fine tuning the existing models.
We conjecture that the visible difference of the patterns in the skeleton images and the images of the natural scenes is the main reason for this phenomenon. 
Hence, it is recommended to train the network from scratch for the full exploitation of the proposed representation.

To demonstrate the additional benefits of augmenting the skeletal image with the speeds of the skeleton joints, we also trained the Inception-ResNet for the augmented images.
Recall, in that case the resulting image has six channels - three channels each for the joint locations and the joint speeds.
To account for the additional information, we modified the Inception-ResNet by extending the ``STEM'' part of the network\cite{szegedy2017inception}. 
The modified architecture is summarized in Fig.~\ref{fig:extended_inception_resnet}.  
To train the modified network, Center loss \cite{wen2016discriminative} is added to the cross entropy to form the final loss function. We optimized the network with the $RMSProp$ optimizer, and selected the initial learning rate as 0.1. 
The results of our experiments (Section~\ref{sec:Exp}) demonstrate a consistent gain in the performance of the network by using the augmented images.

\subsection{Macro-temporal encoding and classification}
\label{sec:TEC}
Once it is possible to process the skeleton data with the desired CNN, it also becomes practicable to exploit the CNN features to further process the skeletal information. 
For instance, as noted in Section~\ref{temporal_extension}, a single skeleton image used in this work represents the temporal information for only $n$ skeletal frames, which  encodes the micro-temporal patterns in an action. To explore the long term temporal relationship of the skeleton joints, we can further perform a macro-temporal encoding over the CNN features. We perform this encoding as follows.

Given a skeleton action video, we first construct the `$Q$' possible skeleton images for the video.
These images are forward passed through the network and the features $\boldsymbol\xi_{i \in \{1,2,...,Q\}} \in \mathbb R^{1792}$ from the {\fontfamily{qcr}\selectfont prelogit} layer of the Inception-Resnet are extracted. 
We compute the Short Fourier Transform~\cite{oppenheim1999discrete} over $\boldsymbol\xi_i, \forall i$ and retain  `$z$' low frequency components of the computed transform.
Next, the column vectors $\boldsymbol\xi_i$ are divided into two equal segments along their row-dimension, and the Fourier Transform is again applied to retain another set of `$z$' low frequency components for each segment.
The procedure is repeated `$\ell$' times and all the $2^{\ell -1} \times z$ resulting components are concatenated to represent the video.
These features are used for training an SVM classifier. We used $\ell = 3$ in our experiments in Section~\ref{sec:Exp}.
The features computed with the above method take into account the whole skeletal sequence in the videos, thereby accounting for the  macro-temporal relations between the skeleton joints.

It is noteworthy that whereas we present the macro-temporal encoding in our approach by employing subsequent processing of the CNN features, the Skepxels-based construction also allows for the direct encoding of the skeletal information over large time intervals using the larger images. 
Nevertheless, in this work, the main objective of the underlying approach  is to demonstrate the effectiveness of the Skepxels-based representation for the common practices of exploiting the CNNs for the skeleton data.
Hence, we intentionally include the explicit processing of the CNN features and show that the state-of-the-art performance is achievable with the compact representations. 



\section{Dataset}
\label{sec:DS}
We perform experiments with three standard benchmark datasets for human action recognition, namely the NTU RGB+D Human Activity Dataset \cite{shahroudy2016ntu}, the Northwestern-UCLA Multiview Dataset~\cite{AOG} and the UTD Multimodal Human Action Dataset~\cite{chen2015utd}.
The details of these datasets and the followed experimental protocols are given below.

\subsection{NTU RGB+D Human Activity Dataset}
\label{sec:NTU}

\begin{figure}[t]
\centering
\includegraphics[width=0.45\textwidth]{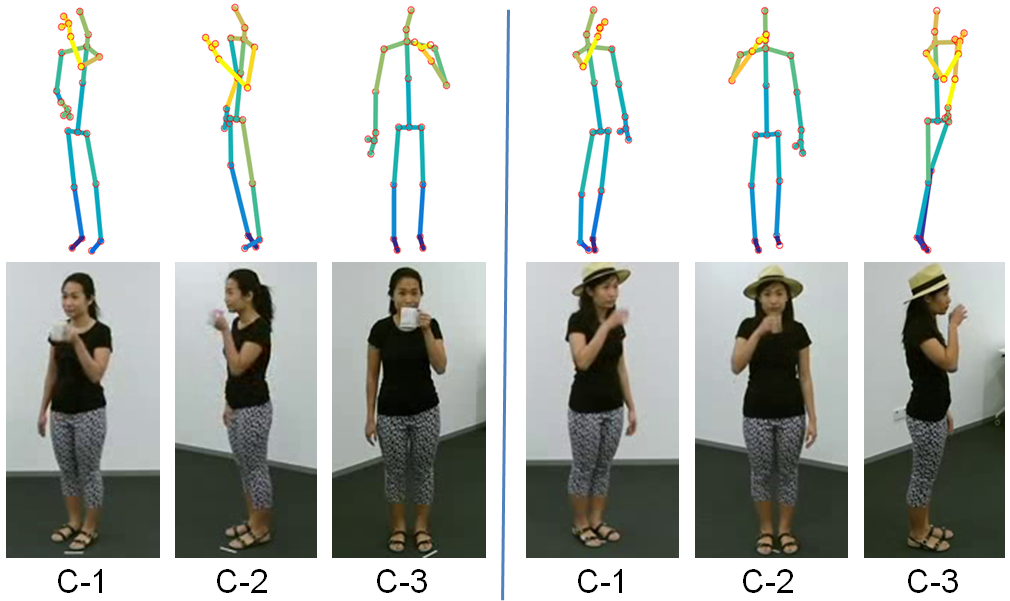}
\centering
\caption{Skeleton and RGB sample frames from the NTU RGB+D Human Activity Dataset \cite{shahroudy2016ntu}. Three sensors C-1, C-2 and C-3 are used for recording. The left image group shows the actions recorded with the performer facing  C-3. The right group is recorded when the action performer faces  C-2.}
\label{fig:ntu_samples}
\end{figure}

The NTU RGB+D Human Activity Dataset \cite{shahroudy2016ntu} is a large-scale RGB+D dataset for human activity analysis. This dataset has been  collected with the Kinect v2 sensor and it includes 56,880 action samples each for RGB, depth, skeleton and infra-red videos. Since we are concerned with the skeleton sequences only, we use the skeleton part of the dataset to evaluate our method. 
In the dataset, there are 40 human subjects performing 60 types of actions including 50 single person actions and 10 two-person interactions. Three sensors were used to capture the data simultaneously from three horizontal angles: $-45^\circ, 0^\circ, 45^\circ$, and every action performer performed the action twice, facing the left or right sensor respectively. Moreover, the height of the sensors and their distances to the action performer have been adjusted in the dataset to get further viewpoint variations. The NTU RGB+D dataset is one of the largest and the most complex cross-view action dataset of its kind to date. Fig.~\ref{fig:ntu_samples} shows representative samples from this dataset.

We followed the standard evaluation protocol proposed in \cite{shahroudy2016ntu}, which includes cross-subject and cross-view evaluations. For the cross-subject case, 40 subjects are equally split into training and testing groups. For the cross-view protocol, the videos captured by the sensor C-2 and C-3 are used as the training samples, whereas the videos captured by the sensor C-1 are used for testing. 


\subsection{Northwestern-UCLA Multiview Dataset}
\label{sec:NUCLA}

\begin{figure}[t]
\centering
\includegraphics[width=0.45\textwidth]{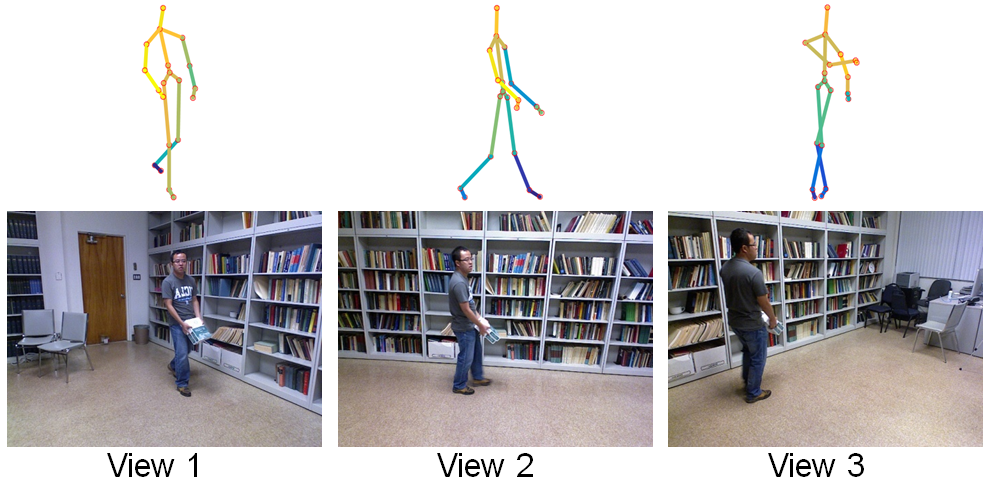}
\caption{Sample frames of three different viewpoints from the NUCLA dataset~\cite{AOG}.}
\label{fig:nucla_samples}
 \end{figure}

This dataset \cite{AOG} contains RGB, Depth and skeleton videos captured simultaneously from three different viewpoints with the Kinect v1 sensor, while we only use skeleton data in our experiments. Fig.~\ref{fig:nucla_samples} shows the representative sample frames from  this dataset for the three viewpoints. The dataset contains videos of 10 subjects performing 10 actions: (1) pick up with one hand, (2) pick up with two hands, (3) drop trash, (4) walk around, (5) sit down, (6) stand up, (7) donning, (8) doffing, (9) throw, and (10) carry. The three viewpoints are: (a) left, (b) front, and (c) right.  This dataset is challenging because some videos share the same ``walking'' pattern before and after the actual action is performed. Moreover, some actions such as ``pick up with on hand'' and ``pick up with two hands'' are hard to distinguish from different viewpoints.

We use skeleton videos captured from two views for training and the third view for testing, which produces three possible cross-view combinations. 

\begin{figure}[t]
\centering
\includegraphics[width=0.45\textwidth]{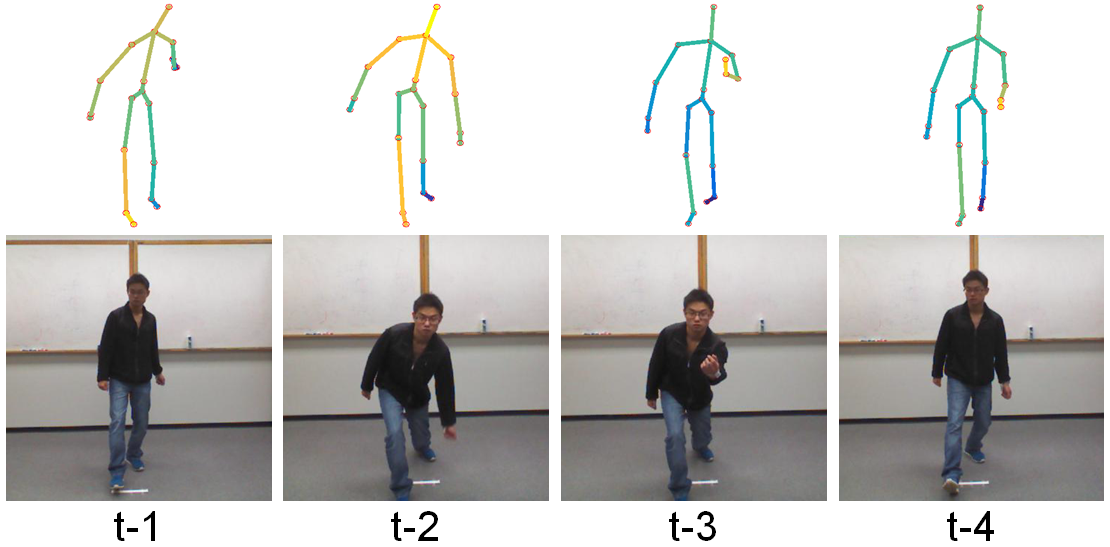}
\caption{Four consecutive sample frames from the UTD-MHAD dataset~\cite{chen2015utd}.}
\label{fig:utd_samples}
 \end{figure}
 

\subsection{UTD Multimodal Human Action Dataset}
\label{sec:UTD}
The UTD-MHAD dataset~\cite{chen2015utd} consists of 27 different actions performed by 8 subjects. Each subject repeated the action for 4 times, resulting in 861 action sequences in total. The RGB, depth, skeleton and the inertial sensor signals were recorded. We only use skeleton videos in our experiments. Fig.~\ref{fig:utd_samples} shows the sample frames from this dataset. We follow \cite{chen2015utd} to evaluation UTD-MHAD dataset with cross-subject protocol, which means the data from subject 1, 3, 5, 7 is used for training, and the data form subject 2, 4, 6, 8 is used for testing.


\section{Experiments}
\label{sec:Exp}

\subsection{Skeleton Image Dimension}
\label{sec:SKID}

\begin{figure}[t]
\centering
\includegraphics[width=0.4\textwidth]{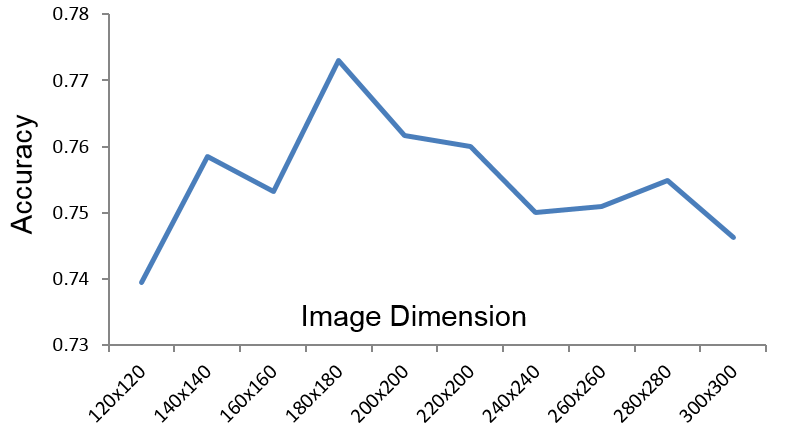}
\caption{Action recognition performance for different skeletal image size  on the NTU RGB+D Human Activity Dataset~\cite{shahroudy2016ntu}. }
\label{fig:image_dimension_compare}
\end{figure}

We first analyze the performance of the models trained with different sizes of the  skeleton images  to choose a suitable image size for our experiments. We used the NTU RGB+D Human Activity Dataset~\cite{shahroudy2016ntu} for this purpose. According to the evaluation protocol of \cite{shahroudy2016ntu}, we split the training samples into training and validation subset. Only the  \emph{location} images were evaluated. After the best image size was chosen, we applied it to both \emph{location} and \emph{velocity} images, and conducted the comprehensive experiments.
During our evaluation for the image size selection, we increased the image size from $120 \times 120$ to $300 \times 300$, with a step of $20$ pixels. Fig.~\ref{fig:image_dimension_compare} shows the recognition accuracy for each setting. We eventually  selected  $180 \times 180$  as the image dimensions based on these results. These skeletal image dimensions were kept the same in our experiments with the other data sets as well.

\subsection{Evaluation on NTU RGB+D Dataset}
We trained our CNN model from scratch for the NTU RGB+D dataset. The model was trained twice for cross-subject and cross-view evaluations respectively. We first evaluated the proposed method with the \emph{location} images only, where the input tensor to the network was in $\mathbb R^{H \times W \times 3}$. We call this evaluation as Skepxel$_{\mathrm{loc}}$ mode. Then, we evaluated our method in Skepxel$_{\mathrm{loc+vel}}$ mode, where we combined the  \emph{location} and \emph{velocity} images to train the network with the  input tensors in $\mathbb R^{H \times W \times 6}$. We used the network defined in Fig.~\ref{fig:extended_inception_resnet} for our evaluation in the Skepxel$_{\mathrm{loc+vel}}$ mode.  Note that some action clips were performed by two persons. In this case we encode each skeleton individually in alternating frames to form a skepxel-based image. This also enable us to readily use the normalization method describe in Section~\ref{sec:NAD}.
Table \ref{tab:ntu_rgbd_comp} compares the performance of our approach with the existing techniques on the NTU dataset. Our method is able to improve the accuracy by $4.4 \%$ in the Skepxel$_{\mathrm{loc+vel}}$ mode over the nearest competitor.

\begin{table}[t]
\centering
\caption{Action recognition accuracy (\%) on the NTU RGB+D Dataset.}
\begin{tabular}{llcc}
\hline\noalign{\smallskip}
\multicolumn{1}{c}{} &  & Cross & Cross \\
Method & Data & Subject & View \\
\noalign{\smallskip}\hline\noalign{\smallskip}
\multicolumn{ 3}{c}{\textbf{Baseline}} &  \\ \hline

Lie Group~\cite{vemulapalli2014human} & Joints & 50.1 & 52.8 \\ 
Deep RNN~\cite{shahroudy2016ntu} & Joints & 56.3 & 64.1 \\ 
HBRNN-L~\cite{du2015hierarchical} & Joints & 59.1 & 64.0 \\ 
Dynamic Skeleton~\cite{hu2015jointly} & Joints & 60.2 & 65.2 \\ 
Deep LSTM~\cite{shahroudy2016ntu} & Joints & 60.7 & 67.3 \\ 
LieNet~\cite{huang2016deep} & Joints & 61.4 & 67.0 \\
P-LSTM~\cite{shahroudy2016ntu} & Joints & 62.9 & 70.3 \\ 
LTMD~\cite{luo2017unsupervised} & Depth & 66.2 & - \\
ST-LSTM~\cite{liu2016spatio} & Joints & 69.2 & 77.7 \\ 
DSSCA-SSLM~\cite{shahroudy2017deep} & RGB-D & 74.9 & - \\ 
Interaction Learning~\cite{rahmani2017learning} & Joints-D & 75.2 & 83.1 \\ 
Clips+CNN+MTLN~\cite{ke2017new} & Joints & 79.6 & 84.8 \\ 

\noalign{\smallskip}\hline\noalign{\smallskip}
\multicolumn{ 3}{c}{\textbf{Proposed}} & \multicolumn{1}{l}{} \\ \hline

Skepxel$_{\mathrm{loc}}$ & Joints & 77.4 & 87.0 \\ 
Skepxel$_{\mathrm{loc+vel}}$ & Joints & \textbf{81.3} & \textbf{89.2} \\ 

\hline\noalign{\smallskip}
\end{tabular}
\label{tab:ntu_rgbd_comp}
\end{table}

\subsection{Evaluation on the NUCLA Dataset}
We took the CNN model trained for the NTU cross-view evaluation as a baseline. Firstly, we directly applied this model on the NUCLA dataset to evaluate the generalization of our model on the unseen skeleton data. Secondly, we fine-tuned the model with the NUCLA dataset and conducted the evaluation again to evaluate  performance on this dataset. 

Table \ref{tab:nucla_comp} summarizes our results on the NUCLA dataset. The proposed method for the skeleton images alone achieves $83.0\%$ average accuracy without fine-tuning on the target dataset, which demonstrates the generalization of our technique.  After  fine-tuning, the average accuracy increases by $2.2\%$. The best performance  is achieved when we combined  the skeleton and the velocity images, improving the  accuracy over the nearest competitor by $5.7\%$.

\begin{table}[t]
\centering
\caption{Action recognition accuracy (\%) on the NUCLA dataset. $V_{1,2}^3$ means that view 1 and 2 were used for training and view 3 was used for testing. Skepxel$_{\mathrm{loc}}^\ast$ used the NTU cross-view model without fine-tuning.}
\begin{tabular}{llcccc}
\hline\noalign{\smallskip}
\multicolumn{ 1}{l}{Method} & \multicolumn{ 1}{l}{Data} & $V_{1,2}^3$ & $V_{1,3}^2$ & $V_{2,3}^1$ & \multicolumn{ 1}{c}{Mean} \\ 
\noalign{\smallskip}\hline\noalign{\smallskip}
\multicolumn{ 6}{c}{\textbf{Baseline}} \\ \hline

Hankelets~\cite{Hankelets} & RGB & 45.2 & - & - & 45.2 \\ 
JOULE~\cite{hu2015jointly} & RGB/D & 70.0 & 44.7 & 33.3 & 49.3 \\ 
DVV~\cite{DVV} & Depth & 58.5 & 55.2 & 39.3 & 51.0 \\ 
CVP~\cite{CVP} & Depth & 60.6 & 55.8 & 39.5 & 52.0 \\ 
AOG~\cite{AOG} & Depth & 73.3 & - & - & - \\ 
nCTE~\cite{nCTE} & RGB & 68.6 & 68.3 & 52.1 & 63.0 \\ 
NKTM~\cite{NKTM} & RGB & 75.8 & 73.3 & 59.1 & 69.4 \\ 
R-NKTM~\cite{rahmani2017pami} & RGB & 78.1 & - & - & - \\ 
HPM+TM~\cite{HPM+TM} & Depth & \textbf{91.9} & 75.2 & 71.9 & 79.7 \\ 

\noalign{\smallskip}\hline\noalign{\smallskip}
\multicolumn{ 6}{c}{\textbf{Proposed}} \\ \hline

Skepxel$_{\mathrm{loc}}^\ast$ & Joints & 89.9 & 83.9 & 75.2 & 83.0 \\ 
Skepxel$_{\mathrm{loc}}$ & Joints & 88.8 & 85.3 & \textbf{81.6} & 85.2 \\ 
Skepxel$_{\mathrm{loc+vel}}$ & Joints & 91.5 & \textbf{85.5} & 79.2 & \textbf{85.4} \\ 

\hline\noalign{\smallskip}
\end{tabular}
\label{tab:nucla_comp}
\end{table}

\begin{table}[t]
\centering
\caption{Action recognition accuracy (\%) on  UTD-MHAD dataset. Skepxel$_{\mathrm{loc}}^\ast$ used the NTU cross-view model without fine-tuning.}
\setlength{\tabcolsep}{15pt}
\vspace{+1mm}
\label{tab:utd_comp}
\begin{tabular}{llc}
\hline\noalign{\smallskip}
\multicolumn{ 1}{l}{Method} & \multicolumn{ 1}{l}{Data} & \multicolumn{ 1}{c}{Mean} \\ 
\noalign{\smallskip}\hline\noalign{\smallskip}
\multicolumn{ 3}{c}{\textbf{Baseline}} \\ \hline

ELC-KSVD~\cite{zhou2014discriminative} & Joints & 76.2 \\ 
kinect-Inertia~\cite{chen2015utd} & Depth & 79.1 \\ 
Cov3DJ~\cite{hussein2013human} & Joints & 85.6 \\ 
SOS~\cite{hou2016skeleton} & Joints & 87.0 \\ 
JTM~\cite{wang2016action} & Joints & 87.9 \\ 

\noalign{\smallskip}\hline\noalign{\smallskip}
\multicolumn{ 3}{c}{\textbf{Proposed}} \\ \hline

Skepxel$_{\mathrm{loc}}^\ast$ & Joints & 94.7 \\ 
Skepxel$_{\mathrm{loc}}$ & Joints & 96.5 \\ 
Skepxel$_{\mathrm{loc+vel}}$ & Joints & {\bf 97.2} \\ 

\hline\noalign{\smallskip}
\end{tabular}
 \end{table}

\subsection{Evaluation on the UTD-MHAD Dataset}
For the UTD-MHAD dataset, we evaluated the performance of our technique using the models pre-trained with the NTU dataset.  The performance of our technique for the different models is summarized in Table~\ref{tab:utd_comp}.  The proposed approach  achieves a significant accuracy gain of 9.3\% on this dataset.






We note that our Skepxels representation can be used with multiple existing CNN architectures, which provides the opportunity to extract varied network features. Exploiting ensembles/concatenation of such features, it is possible to achieve further performance gain using our method. 
We provide discussion on this aspect of our approach in the supplementary material of the paper.

\begin{figure*}[t]
\centering
\includegraphics[width=1\textwidth]{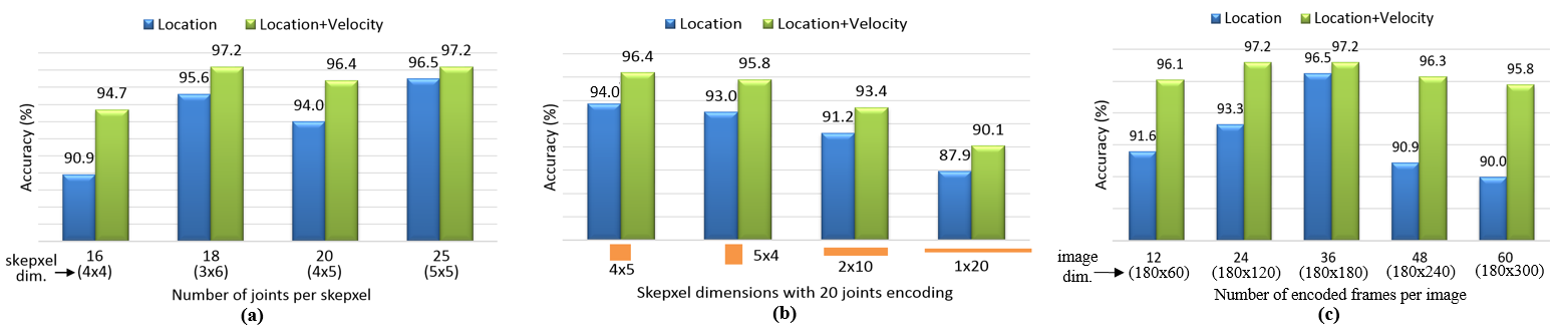}
\caption{Performance on UTD-MHAD dataset~\cite{chen2015utd} with different parameter choices: In (a), the number of joints are  varied by dropping/interpolating joints. In (b), the skepxel dimensions are varied. In (a) and (b) both, the number of skepxels and frames are chosen such that the final image size is  $180\times 180$. For (c), 36 skepxels of size $5\times 5$   are  used per frame, and the number of frames is varied. The resulting images are then resized to $180\times 180$.}
\label{fig:ablation_a}
\end{figure*}

\section{Ablation Experiments}
\label{sec:AbS}

\subsection{Testing Different Skepxel Encoding Schemes}
We conducted this ablation study to demonstrate how the overall recognition performance is affected by (a)~altering the number of joints encoded per Skepxel, (b)~changing the Skepxel dimensions with fixed number of joints; and (c)~changing the number of frames encoded per image. We chose UTD-MHAD dataset~\cite{chen2015utd} for these experiments because the proposed representation achieved significant performance improvements for this dataset. The summary of the results of our ablation experiments is presented in Fig.~\ref{fig:ablation_a}. The overall results demonstrate effective encoding by skepxels. The location+velocity encoding is always able to improve the performance, which is intuitive because of more information being encoded in the representation.

\subsection{Testing Different CNN Architectures}
\label{sec:architectures}
Results in the previous Sections suggest that Skepxel is an effective representation for encoding human skeletons as it allows CNNs to exploit both spatial-only and spatio-temporal convolutions.
It is also flexible in generating images of arbitrary dimensions to suit different CNN architectures. 
Hence, we also tested Skepxels with two additional CNN architectures, GoogLeNet~\cite{GoogLeNet} and ResNet~\cite{ResNet} to demonstrate its generalization across different models. We first trained GoogLeNet and ResNet using the NTU RGB+D Dataset~\cite{shahroudy2016ntu} from scratch. The same normalization and data augmentation strategies were adopted in these  experiments as described in Section~\ref{sec:NAD}. The trained models were then tested on the NUCLA Dataset~\cite{AOG} and UTD-MHAD Dataset~\cite{chen2015utd}. For both networks, we extracted features from the layers prior to the {\fontfamily{qcr}\selectfont softmax} layer resulting in $\boldsymbol\xi_{i \in \{1,2,...,Q\}} \in \mathbb R^{1024}$ and $\boldsymbol\xi_{i \in \{1,2,...,Q\}} \in \mathbb R^{2048}$ features for GoogLeNet and ResNet, respectively. Results of our experiments are reported in Table~\ref{tab:nucla_comp2} and \ref{tab:utd_comp2}. The results also include the performance of Inception-ResNet from  Section~\ref{sec:Exp} for comparison. Tables~\ref{tab:nucla_comp2} and \ref{tab:utd_comp2} depict good generalization of the proposed representation across different network models. 

\begin{table*}[t]
\centering
\caption{Action recognition accuracy (\%) of different models using the proposed representation on the NUCLA dataset. $V_{1,2}^3$ means that view 1 and 2 were used for training and view 3 was used for testing. Skepxel$_{\mathrm{loc + vel}}$ means CNN model trained with both \emph{location} and \emph{velocity} images. Skepxel$_{\mathrm{loc}}$ only uses \emph{location} images.}
\setlength{\tabcolsep}{5pt}
\begin{tabular}{lccccc}
\hline\noalign{\smallskip}
\multicolumn{ 1}{l}{Method} & \multicolumn{ 1}{c}{CNN} & $V_{1,2}^3$ & $V_{1,3}^2$ & $V_{2,3}^1$ & \multicolumn{ 1}{c}{Mean} \\ 

\noalign{\smallskip}\hline\noalign{\smallskip}
\multicolumn{ 6}{c}{\textbf{Single Network}} \\ 
\noalign{\smallskip}\hline\noalign{\smallskip}

Skepxel$_{\mathrm{loc}}$ & Inception-ResNet & 88.8 & 85.3 & 81.6 & 85.2 \\ 
\noalign{\smallskip}
Skepxel$_{\mathrm{loc+vel}}$ & Inception-ResNet & 91.5 & 85.5 & 79.2 & 85.4 \\ 
\noalign{\smallskip}
Skepxel$_{\mathrm{loc}}$ & GoogLeNet& 89.9 & 82.5 & 74.9 & 82.4 \\ 
\noalign{\smallskip}
Skepxel$_{\mathrm{loc+vel}}$ & GoogLeNet & 91.0 & 86.1 & 76.0 & 84.4 \\ 
\noalign{\smallskip}
Skepxel$_{\mathrm{loc}}$ & ResNet & 88.4 & 83.7 & 78.6 & 83.6 \\ 
\noalign{\smallskip}
Skepxel$_{\mathrm{loc+vel}}$ & ResNet & 91.7 & 85.7 & 80.8 & 86.1 \\ 
\noalign{\smallskip}

\noalign{\smallskip}\hline\noalign{\smallskip}
\multicolumn{ 6}{c}{\textbf{Network Ensemble}} \\ 
\noalign{\smallskip}\hline\noalign{\smallskip}

Skepxel$_{\mathrm{loc}}$ & Inception-ResNet+GoogLeNet & 92.1 & 86.1 & 82.6 & 86.9 \\ 
\noalign{\smallskip}
Skepxel$_{\mathrm{loc+vel}}$ & Inception-ResNet+GoogLeNet & 94.1 & 86.7 & 83.0 & 87.9 \\ 
\noalign{\smallskip}
Skepxel$_{\mathrm{loc}}$ & Inception-ResNet+ResNet & 91.9 & 86.9 & 80.2 & 86.3 \\ 
\noalign{\smallskip}
Skepxel$_{\mathrm{loc+vel}}$ & Inception-ResNet+ResNet & 93.9 & 87.3 & 83.8 &	88.3 \\ 
\noalign{\smallskip}

\hline\noalign{\smallskip}
\end{tabular}
\label{tab:nucla_comp2}
\end{table*}

\begin{table}[t]
\centering
\caption{Action recognition accuracy (\%) of different models using the proposed representation on the UTD-MHAD dataset. }
\setlength{\tabcolsep}{5pt}
\begin{tabular}{lcc}
\hline\noalign{\smallskip}
\multicolumn{ 1}{l}{Method} & \multicolumn{ 1}{c}{CNN} & \multicolumn{ 1}{c}{Mean} \\ 

\noalign{\smallskip}\hline\noalign{\smallskip}
\multicolumn{ 3}{c}{\textbf{Single Network}} \\ 
\noalign{\smallskip}\hline\noalign{\smallskip}

Skepxel$_{\mathrm{loc}}$ & Inception-ResNet & 96.5 \\ 
\noalign{\smallskip}
Skepxel$_{\mathrm{loc+vel}}$ & Inception-ResNet & 97.2 \\ 
\noalign{\smallskip}
Skepxel$_{\mathrm{loc}}$ & GoogLeNet & 88.8 \\ 
\noalign{\smallskip}
Skepxel$_{\mathrm{loc+vel}}$ & GoogLeNet & 94.9 \\ 
\noalign{\smallskip}
Skepxel$_{\mathrm{loc}}$ & ResNet & 88.6 \\ 
\noalign{\smallskip}
Skepxel$_{\mathrm{loc+vel}}$ & ResNet & 92.6 \\ 
\noalign{\smallskip}

\noalign{\smallskip}\hline\noalign{\smallskip}
\multicolumn{ 3}{c}{\textbf{Network Ensemble}} \\ 
\noalign{\smallskip}\hline\noalign{\smallskip}

Skepxel$_{\mathrm{loc}}$ & Inception-ResNet+GoogLeNet & 96.7 \\ 
\noalign{\smallskip}
Skepxel$_{\mathrm{loc+vel}}$ & Inception-ResNet+GoogLeNet & 98.6 \\ 
\noalign{\smallskip}
Skepxel$_{\mathrm{loc}}$ & Inception-ResNet+ResNet & 94.2 \\ 
\noalign{\smallskip}
Skepxel$_{\mathrm{loc+vel}}$ & Inception-ResNet+ResNet & 97.7 \\ 
\noalign{\smallskip}

\hline\noalign{\smallskip}
\end{tabular}
\label{tab:utd_comp2}
\end{table}

\subsection{Using an Ensemble of CNN Models}
\label{sec:ensembling}
The possibility of using Skepxels with multiple CNN architectures also allows us to exploit ensemble of networks to improve  recognition performance.
Therefore, we also performed experiments using network ensembles. To that end, we concatenated the features learned from multiple networks and fed them to our classification scheme. We also performed these experiments with NUCLA and UTD-MHAD Datasets.
The results of our analysis are also summarized in Tables~\ref{tab:nucla_comp2} and \ref{tab:utd_comp2}. Improvements in the recognition performances using the ensembles of features under the Skepxels representation demonstrate its potential for contributing towards advancing the action recognition accuracy with the ever improving network architectures in the future.


\subsection{Testing Skepxels on RGB Actions}
To demonstrate that the effectiveness of Skepxels is not limited to the datasets that provide precise skeleton information, we also performed experiments by extracting inaccurate skeleton information from RGB vidoes, and feeding the resulting skeltons to our approach.
We extracted the skeletons from UTD-MHAD dataset~\cite{chen2015utd} using the DeeperCut method~\cite{insafutdinov2016deepercut} that gives 14 joints per frame. We added 2 more joints by interpolating between left/right hip, and hip/neck and formed images with $4\times 4$ skepxels, assigning zeros to the z-axis values. With this setting, the recognition accuracies for the UTD-MHAD dataset are 87.2\% and 92.3\% for \emph{loc.} and \emph{loc.+vel.}, respectively. These results are comparable to the first two bars in Fig.~\ref{fig:ablation_a} that are correspondingly computed for $4\times4$ skepxels with the provided accurate 3d skeletons. This demonstrates that the proposed representation is almost as effective for recognizing actions from RGB videos as it is from the precise skeleton sequences.



\section{Conclusion}
\label{sec:Conc}
This article proposed a novel method to map skeletal data to images that are effectively processed by CNN architectures.
The method exploits a basic building block, termed \emph{Skepxel} - skeleton picture element, to construct the skeletal images of arbitrary dimensions. The resulting images encode fine spatio-temporal information about the human skeleton under multiple informative joint arrangements in different frames of the skeleton videos. 
This representation is further extended to incorporate the joint speed information in the videos.
Moreover, it is also shown that the proposed compact representation can be easily used  to successfully capture the macro-temporal details in the videos.
When used with the Inception-ResNet architecture, the proposed skeletal image representation result in the state-of-the-art skeleton action recognition performance on the  standard large scale action recognition datasets. 
Further ablation study shows that (1) the proposed Skepxel representation generalizes well over differnt CNN architectures,  (2) the recognition performance of the representation improves with better networks, and (3) an ensemble of CNNs improves the accuracy further; nevertheless, the performance also remains comparable when applied to individual networks. This shows that the representation is sufficient for extracting features using a single CNN model which is important as it save computation time and memory requirements.

\section*{Acknowledgment}
This research was supported by ARC grant DP160101458. The Titan Xp used for this research was donated by NVIDIA Corporation.

\ifCLASSOPTIONcaptionsoff
  \newpage
\fi



\bibliographystyle{IEEEtran}
\bibliography{skepxel_IEEE}
%



%

\begin{IEEEbiography}[{\includegraphics[width=1in,height=1.25in,clip,keepaspectratio]{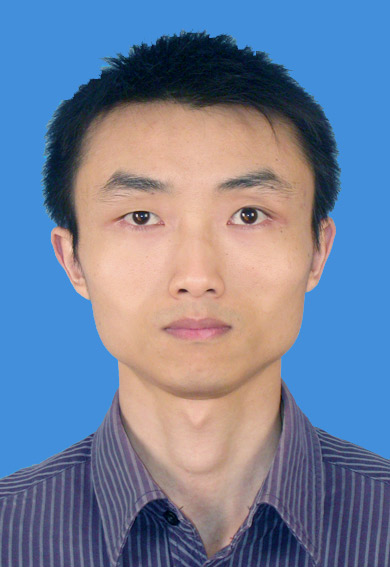}}]{Jian Liu} 
is currently working towards his PhD degree at the School of Computer Science and Software Engineering, The University of Western Australia.
He received his M.S. degree in 2011 from The Hongkong University of Science and Technology, and received his B.E. degree in 2006 from Huazhong University of Science and Technology.  His research interests include computer vision, human action recognition, deep learning, and human pose estimation. His work in these areas have been published in top-ranked research venues, including IEEE CVPR, IEEE TIP and IJCV. 

\end{IEEEbiography}

\begin{IEEEbiography}[{\includegraphics[width=1in,height=1.25in,clip,keepaspectratio]{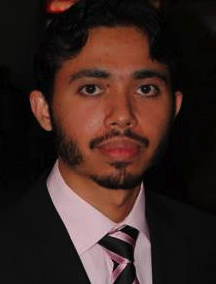}}]{Naveed Akhtar} 
received his PhD in Computer Vision from The University of Western Australia (UWA) and Master degree in Autonomous Systems from Hochschule Bonn-Rhein-Sieg, Germany (HBRS). His research in  Computer Vision and Pattern Recognition has regularly published in prestigious venues of the field, including IEEE CVPR and IEEE TPAMI. He has also served as a reviewer for these venues. During his PhD, he was the recipient of multiple scholarships, and winner of the Canon Extreme Imaging Competition in 2015. Currently, he is a Research Fellow at UWA since July 2017. Previously, he has also served on the same position at the Australian National University. His current research interests include machine learning, action recognition and hyperspectral image analysis.

\end{IEEEbiography}

\begin{IEEEbiography}[{\includegraphics[width=1in,height=1.25in,clip,keepaspectratio]{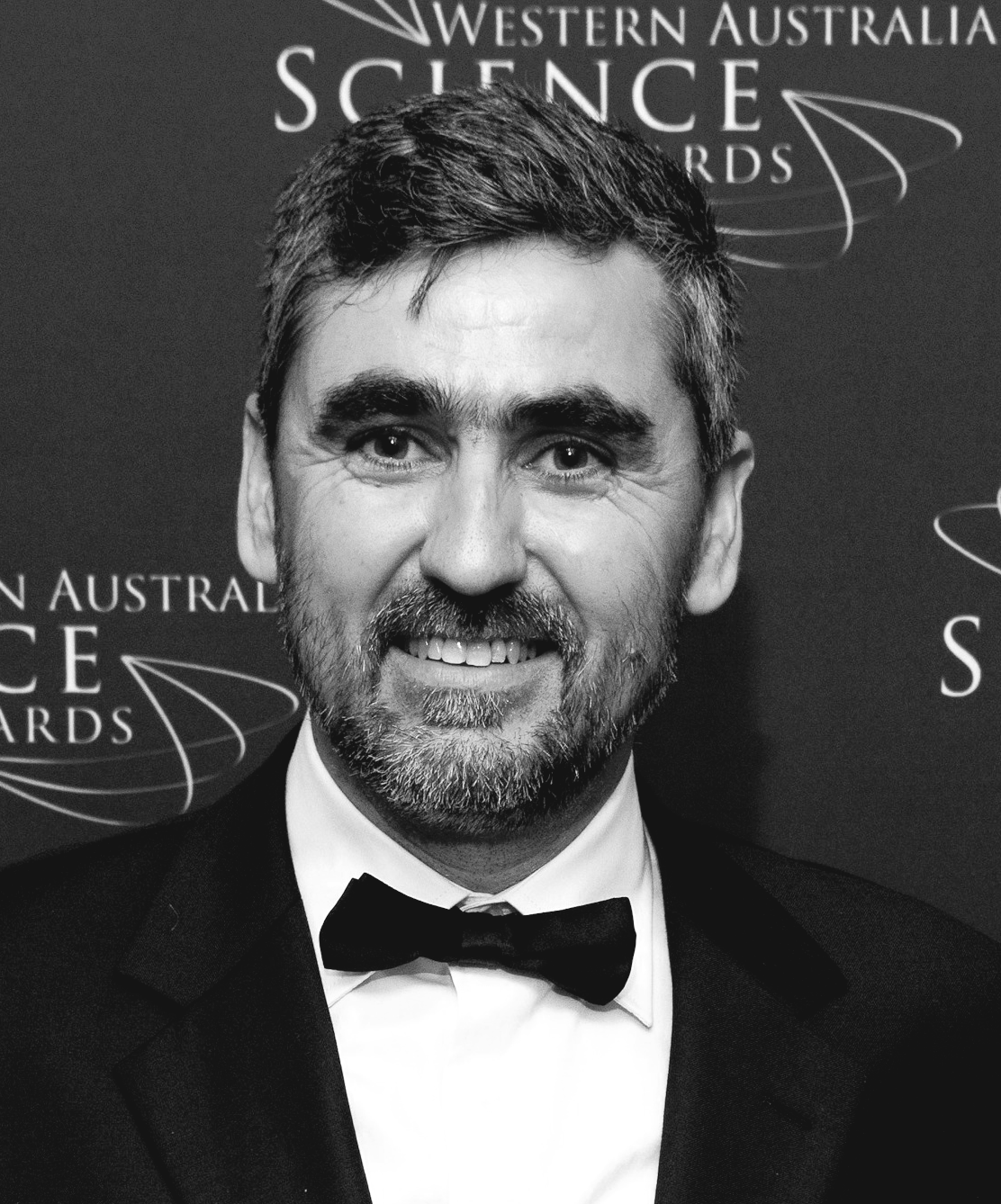}}]{Ajmal Mian} completed his PhD from The University of Western Australia in 2006 with distinction and received the Australasian Distinguished Doctoral Dissertation Award from Computing Research and Education Association of Australasia. He received the prestigious Australian Postdoctoral and Australian Research Fellowships in 2008 and 2011 respectively. He received the UWA Outstanding Young Investigator Award in 2011, the West Australian Early Career Scientist of the Year award in 2012 and the Vice-Chancellors Mid-Career Research Award in 2014. He has secured seven Australian Research Council grants and one National Health and Medical Research Council grant with a total funding of over \$3 Million. He is currently in the School of Computer Science and Software Engineering at The University of Western Australia and is a guest editor of Pattern Recognition, Computer Vision and Image Understanding and Image and Vision Computing journals. His research interests include computer vision, machine learning, 3D shape analysis, hyperspectral image analysis, pattern recognition, and multimodal biometrics.
\end{IEEEbiography}



\end{document}